\def\year{2017}\relax
\documentclass[letterpaper]{article}
\usepackage{aaai17}
\usepackage{times}
\usepackage{helvet}
\usepackage{courier}
\usepackage{amsmath}
\usepackage{graphicx}
\usepackage{epstopdf}
\usepackage{multirow}
\usepackage{makecell}
\usepackage{url}
\usepackage{multirow}
\usepackage{makecell}
\begin{document}
%
\title{Topic Aware Neural Response Generation}
\author{
	Chen Xing$^1$ $^2$~~~, Wei Wu$^4$~~~, Yu Wu$^3$~~~~, Jie Liu$^1$ $^2$~~~,\\
	\textbf{Yalou Huang}$^1$ $^2$~~~~, \textbf{Ming Zhou}$^4$~~~~, \textbf{Wei-Ying Ma}$^4$~~~~\\
	$^1$College of Computer and Control Engineering, Nankai University, Tianjin, China\\
	$^2$College of Software, Nankai University, Tianjin, China\\
	$^3$State Key Lab of Software Development Environment, Beihang University, Beijing, China\\
	$^4$Microsoft Research, Beijing, China} 

\maketitle
	
\begin{abstract}
We consider incorporating topic information into the sequence-to-sequence framework to generate informative and interesting responses for chatbots. To this end, we propose a topic aware sequence-to-sequence (TA-Seq2Seq) model.  The model utilizes topics to simulate prior knowledge of human that guides them to form informative and interesting responses in conversation, and leverages the topic information in generation by a joint attention mechanism and a biased generation probability. The joint attention mechanism summarizes the hidden vectors of an input message as context vectors by message attention, synthesizes topic vectors by topic attention from the topic words of the message obtained from a pre-trained LDA model, and let these vectors jointly affect the generation of words in decoding. To increase the possibility of topic words appearing in responses, the model modifies the generation probability of topic words by adding an extra probability item to bias the overall distribution. Empirical study on both automatic evaluation metrics and human annotations shows that TA-Seq2Seq can generate more informative and interesting responses, and significantly outperform the-state-of-the-art response generation models.
\end{abstract}

\section{Introduction}
Human-computer conversation is a challenging task in AI and NLP. Existing conversation systems include task oriented dialog systems \cite{young2013pomdp} and non task oriented chatbots. Dialog systems aim to help people complete specific tasks such as ordering and tutoring, while chatbots are designed for realizing natural and human-like conversation with people regarding to a wide range of issues in open domains \cite{perez2011conversational}. Although previous research focused on dialog systems, recently, with the large amount of conversation data available on the Internet, chatbots are becoming hot in both academia and industry.  

A common approach to building the conversation engine in a chatbot is learning a response generation model within a machine translation (MT) framework  \cite{ritter2011data,sutskever2014sequence,shang2015neural,DBLP:conf/naacl/SordoniGABJMNGD15} from the large scale social conversation data. Recently, neural network based methods have become the mainstream because of their capability to capture semantic and syntactic relations between messages and responses in a scalable and end-to-end way.  Sequence-to-sequence (Seq2Seq) with attention \cite{bahdanau2014neural,cho2015describing} represents the state-of-the-art neural network model for response generation. To engage people in conversation, the response generation algorithm in a chatbot should generate responses that are not only natural and fluent, but also informative and interesting.  MT models such as Seq2Seq with attention, however, tend to generate trivial responses like ``me too'', ``I see'', or ``I don't know'' \cite{li2015diversity} due to the high frequency of these patterns in data. Although these responses are safe to reply to many messages, they are boring and carry little information. Such responses may quickly lead the conversation between human and machine to an end, and severely hurt the user experience of a chatbot.  

In this paper, we study the problem of response generation for chatbots. Particularly, we target to generate informative and interesting responses that can help chatbots engage their users. Unlike Li et al. \cite{li2015diversity} who tried to passively avoid generating trivial responses by penalizing their generation probabilities, we consider solving the problem by actively bringing content into responses by topics.  Given an input message, we predict possible topics that can be talked about in responses, and generate responses with the topics. The idea is inspired by our observation on conversation between humans. In human-human conversation, people often associate an input message with topically related concepts in their mind. Based on the concepts, they organize content and select words for their responses. For example, to reply to ``my skin is so dry'', people may think it is a ``skin'' problem and can be alleviated by ``hydrating'' and ``moisturizing''.  Based on this knowledge, they may give more informative responses like ``then hydrate and moisturize our skin'' rather than trivial responses like ``me too''.  The informative responses could let other people follow the topics and continue talking about skin care.  ``Skin'', ``hydrate'', and ``moisturize'' are topical concepts related to the message. They represent people's prior knowledge in conversation.  In responding, people will bring content that are relevant to the concepts to their responses and even directly use the concepts as building blocks to form their responses.  

We consider simulating the way people respond to messages with topics, and propose a topic aware sequence-to-sequence (TA-Seq2Seq) model in order to leverage topic information as prior knowledge in response generation. TA-Seq2Seq is built on the sequence-to-sequence framework. In encoding, the model represents an input message as hidden vectors by a message encoder, and acquires embeddings of the topic words of the message from a pre-trained Twitter LDA model. The topic words are used as a simulation of topical concepts in people's mind, and obtained from a Twitter LDA model which is pre-trained using large scale social media data outside the conversation data. In decoding, each word is generated according to both the message and the topics through a joint attention mechanism. In joint attention, hidden vectors of the message are summarized as context vectors by message attention  which follows the existing attention techniques, and embeddings of topic words are synthesized as topic vectors by topic attention. Different from the existing attention, in topic attention, the weights of the topic words are calculated by taking the final state of the message as an extra input in order to strengthen the effect of the topic words relevant to the message. The joint attention lets the context vectors and the topic vectors jointly affect response generation, and makes words in responses not only relevant to the input message, but also relevant to the correlated topic information of the message.  To model the behavior that people use topical concepts as ``building blocks'' of their responses, we modify the generation probability of a topic word by adding another probability item which biases the overall distribution and further increases the possibility of the topic word appearing in the response.

We conduct empirical study on large scale data crawled from Baidu Tieba, 
and compare different methods by both automatic evaluation and human judgment. The results on both automatic evaluation metrics and human annotations show that TA-Seq2Seq can generate more informative, diverse, and topic relevant responses and significantly outperform the-state-of-the-art methods for response generation.    

The contributions of this paper include 1) proposal of using topics as prior knowledge for response generation; 2) proposal of a TA-Seq2Seq model that naturally incorporates topic information into the encoder-decoder structure; 3) empirical verification of the effectiveness of TA-Seq2Seq.

\section{Background: sequence-to-sequence model and attention mechanism}
\label{background}
Before introducing our model, let us first briefly review the Seq2Seq model and the attention mechanism.

\subsection{Sequence-to-sequence model}
In Seq2Seq, given a source sequence (message) $\mathbf{X}=(x_1,x_2,\ldots,x_T)$ and a target sequence (response) $\mathbf{Y}=(y_1,y_2,\ldots,y_{T'})$, the model maximizes the generation probability of $\mathbf{Y}$ conditioned on  $\mathbf{X}$: $p(y_1,...,y_{T'}|x_1,...,x_T)$. Specifically, Seq2Seq  is in an encoder-decoder structure. The encoder reads $\mathbf{X}$ word by word and represents it as a context vector $\mathbf{c}$ through a recurrent neural network (RNN), and then the decoder estimates the generation probability of $\mathbf{Y}$ with $\mathbf{c}$ as input.  The objective function of Seq2Seq can be written as
\begin{equation*}
\label{eq: objective function of s2s}
p(y_1,...,y_{T'}|x_1,...,x_T)=p(y_1|\mathbf{c})\prod_{t=2}^{T'}p(y_t|\mathbf{c},y_1,...,y_{t-1}).
\end{equation*}
The encoder RNN calculates the context vector $\mathbf{c}$ by
\begin{equation*}
\label{eq: encoder function}
\mathbf{h}_t=f(x_t,\mathbf{h}_{t-1}); \mathbf{c}=\mathbf{h}_{T},
\end{equation*}
where $\mathbf{h}_t$ is the hidden state at time $t$ and $f$ is a non-linear transformation which can be either an long-short term memory unit (LSTM) \cite{hochreiter1997long} or a gated recurrent unit (GRU) \cite{cho2014properties}. In this work, we implement $f$ using GRU which is parameterized as
\begin{equation}
\label{GRU unit}
\begin{aligned}
&\mathbf{z}=\sigma(\mathbf{W}^z\mathbf{x}_t+\mathbf{U}^z\mathbf{h}_{t-1})\\
&\mathbf{r}=\sigma(\mathbf{W}^r\mathbf{x}_t+\mathbf{U}^r\mathbf{h}_{t-1})\\
&\mathbf{s}=tanh(\mathbf{W}^s\mathbf{x}_t+\mathbf{U}^s(\mathbf{h}_{t-1}\circ \mathbf{r}))\\
&\mathbf{h}_t=(1-\mathbf{z})\circ \mathbf{s}+\mathbf{z}\circ\mathbf{h}_{t-1}
\end{aligned}
\end{equation}
The decoder is a standard RNN language model except conditioned on the context vector $\mathbf{c}$. The probability distribution $\mathbf{p}_t$ of candidate words at every time $t$ is calculated as
\begin{equation*}
\label{eq: decoder function}
\mathbf{s}_t=f(y_{t-1},\mathbf{s}_{t-1},\mathbf{c}); \mathbf{p}_t=softmax(\mathbf{s}_t,y_{t-1})
\end{equation*}
where $\mathbf{s}_t$ is the hidden state of the decoder RNN at time $t$ and $y_{t-1}$ is the word at time $t-1$ in the response sequence.

\subsection{Attention mechanism}
The traditional Seq2Seq model assumes that every word is generated from the same context vector. In practice, however, different words in $\mathbf{Y}$ could be semantically related to different parts of $\mathbf{X}$.  To tackle this issue,  attention mechanism \cite{bahdanau2014neural} is introduced to Seq2Seq. In Seq2Seq with attention, each $y_i$ in $\mathbf{Y}$ corresponds to a context vector $\mathbf{c}_i$, and $\mathbf{c}_i$ is a weighted average of all hidden states $\{ \mathbf{h}_t \}_{t=1}^T$ of the encoder. Formally,  $\mathbf{c}_i$ is defined as
\begin{equation}
\label{eq: attention-weighted sum}
\mathbf{c}_i=\Sigma_{j=1}^T\alpha_{ij}\mathbf{h}_j,
\end{equation}
where $\alpha_{ij}$ is given by
\begin{equation}
\label{eq: attention-normalize}
\begin{aligned}
\alpha_{ij}=\frac{exp(e_{ij})}{\Sigma_{k=1}^T exp(e_{ik})}; e_{ij}=\eta(\mathbf{s}_{i-1},\mathbf{h}_j)
\end{aligned}
\end{equation}
$\eta$ is usually implemented as a multi-layer perceptron (MLP) with tanh as an activation function.
\begin{figure}[t]
\begin{center}
\includegraphics[width=5.5cm,height=3cm]{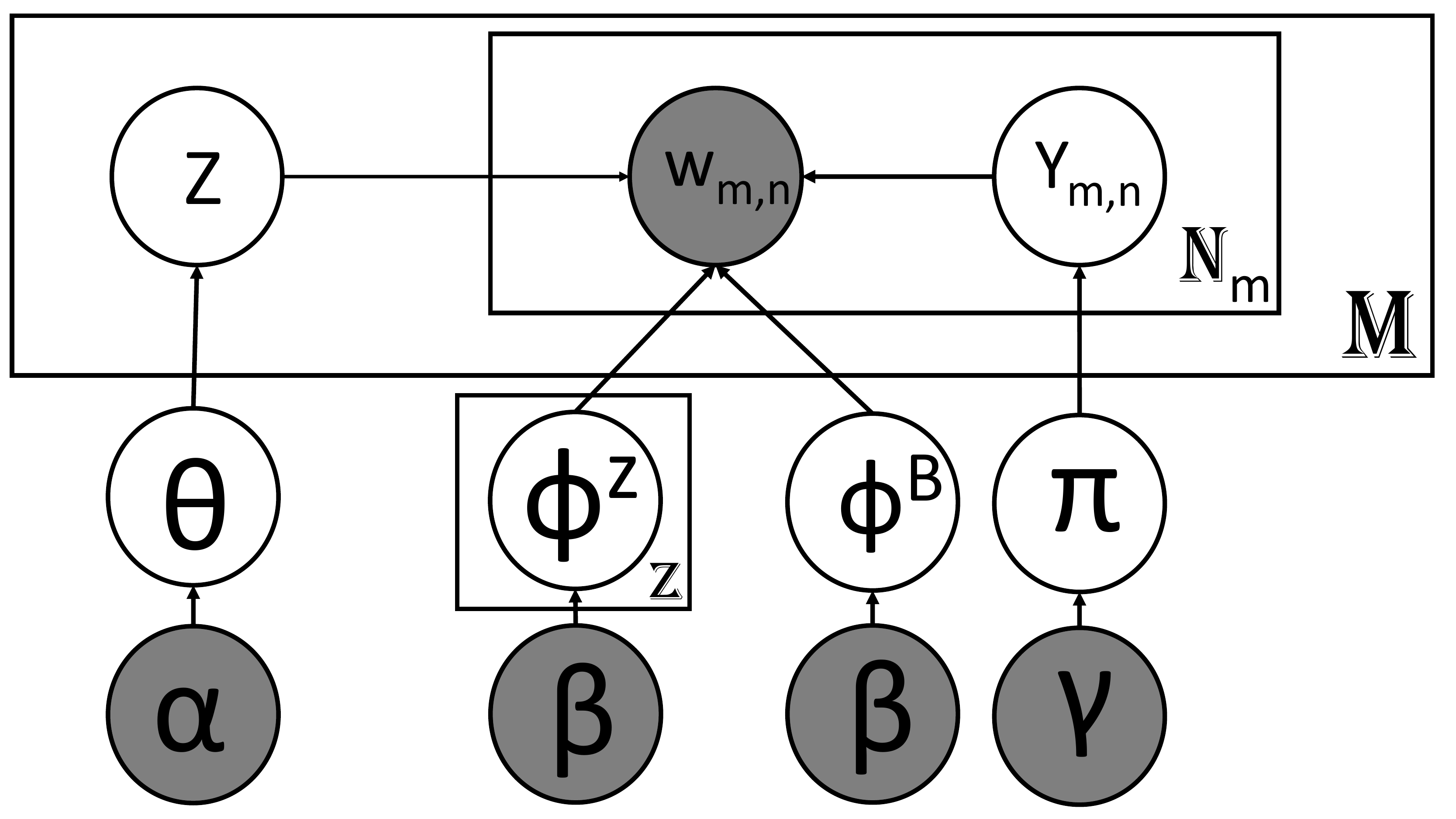}
\end{center}
\caption{Graphic model of Twitter LDA}\label{fig:topic_model}
\end{figure}
\section{Topic aware Seq2Seq model}
Suppose that we have a data set $\mathcal{D}=\{(\mathbf{K}_i,\mathbf{X}_i,\mathbf{Y}_i)\}_{i=1}^N$ where $\mathbf{X}_i$ is a message, $\mathbf{Y}_i$ is a response, and $\mathbf{K}_i=(\mathbf{k}_{i,1},\ldots,\mathbf{k}_{i,n})$ are the topic words of $\mathbf{X}_i$.  
Our goal is to learn a response generation model from $\mathcal{D}$, and thus given a new message $\mathbf{X}$ with topic words $\mathbf{K}$, the model can generate response candidates for $\mathbf{X}$.   
 
To learn the model, we need to answer two questions: 1) how to obtain the topic words; 2) how to perform learning. 
In this section, we first describe our method on topic word acquisition, and then we give details of our model.  
\begin{figure*}[t]
\begin{center}
\includegraphics[width=16cm,height=8cm]{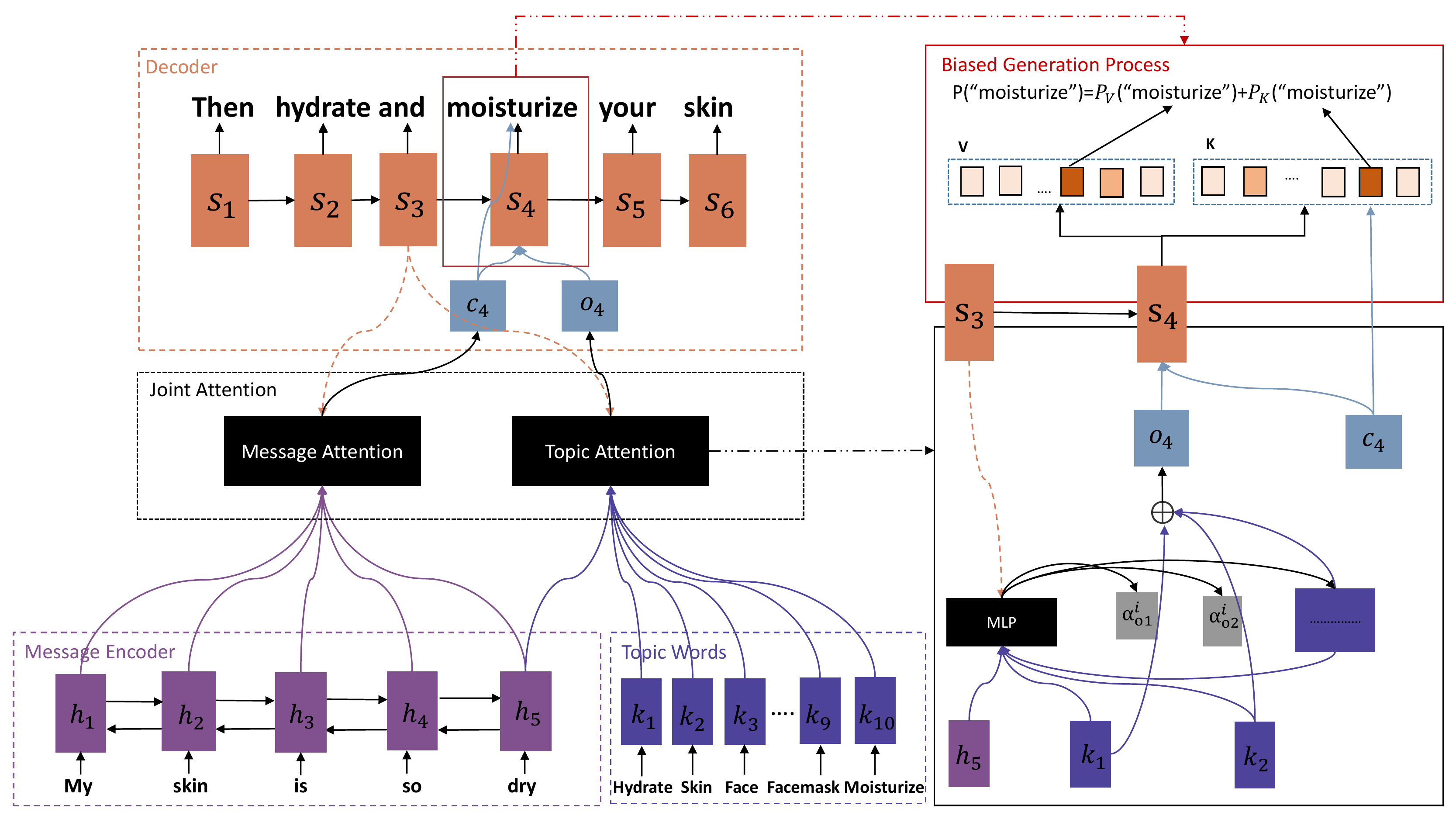}
\end{center}
\caption{Structure of TA-Seq2Seq}\label{fig:general_model}
\end{figure*}
\subsection{Topic word acquisition}
We obtain topic words of a message from a Twitter LDA model \cite{zhao2011comparing}. Twitter LDA belongs to the family of probabilistic topic models \cite{blei2003latent} and represents the state-of-the-art topic model for short texts \cite{zhao2011comparing}.  The basic assumption of Twitter LDA is that each message corresponds to one topic, and each word in the message is either a background word or a topic word under the topic of the message. Figure \ref{fig:topic_model} gives the graphical model of Twitter LDA.

We estimate the parameters of Twitter LDA using the collapsed Gibbs sampling algorithm \cite{zhao2011comparing}. After that, we use the model to assign a topic $z$ to a message $\mathbf{X}$,  pick the top $n$ words ($n=100$ in our experiments) with the highest probabilities under $z$, and remove universal words like ``thank'' and ``you'' to get the topic words $\mathbf{K}$ for $\mathbf{X}$.  

In learning, we need a vector representation for each topic word. To this end, we first calculate a distribution for topic word $w$  by Equation (\ref{scoreword}) where $C_{wz}$ is the number of times that $w$ is assigned to topic $z$ in training. Then, we take the distributions as the vector representations of the topic words. 
\begin{equation} \label{scoreword}
p(z|w) \propto \frac{C_{wz}}{\sum_{z'}C_{ wz'}}.
\end{equation}

In our experiments, we trained a Twitter LDA model using large scale posts from Sina Weibo which is the largest microblogging service in China. The data provides topic knowledge apart from that in message-response pairs that we use to train the response generation model. The process is similar to how people learn to respond in conversation: they become aware of what can be talked about from Internet, especially from social media, and then use what they learned as topics to form their responses in conversation.  	
	
Note that in addition to LDA, one can employ other techniques like tag recommendation \cite{wu2016improving} or keyword extraction \cite{wu2015mining} to generate topic words. One can also get topic words from other resources like wikipedia and other web documents. We leave the discussion of these extensions as our future work. 
\subsection{Model}
Figure \ref{fig:general_model} gives the structure of topic aware sequence-to-sequence model (TA-Seq2Seq). TA-Seq2Seq is built on the sequence-to sequence framework, and leverages topic information by a joint attention mechanism and a biased generation probability.  

Specifically, in encoding, a message encoder represents an input message $\mathbf{X}$ 
as a series of hidden vectors $\{\mathbf{h}_t\}_{t=1}^T$ by a bidirectional GRU-RNN from both ends\footnote{Hidden vectors from both directions are concatenated together.}. GRU is defined in Equation (\ref{GRU unit}). At the same time, a topic encoder obtains the embeddings of the topic words $\mathbf{K}$ of $\mathbf{X}$ by looking up an embedding table which is established according to Equation (\ref{scoreword}). With a little abuse of notations, we also use $(\mathbf{k}_1,\ldots,\mathbf{k}_n)$ to denote the the embeddings of words in $\mathbf{K}$. The meaning of $(\mathbf{k}_1,\ldots,\mathbf{k}_n)$ is clear in context. 
 
In decoding, at step $i$, message vectors $\{\mathbf{h}_t\}_{t=1}^T$ are transformed to a context vector $\mathbf{c}_i$ by message attention given by Equation (\ref{eq: attention-weighted sum}) and Equation (\ref{eq: attention-normalize}), and embeddings of topic words $\{\mathbf{k}_j\}_{j=1}^n$ are linearly combined as a topic vector $\mathbf{o}_i$ by topic attention. The combination weight of $\mathbf{k}_j$ is given by
\begin{equation}
\label{eq: topical-attention-weighted sum}
\alpha^i_{oj}=\frac{exp(\eta_o(\mathbf{s}_{i-1},\mathbf{k}_j,\mathbf{h}_T))}{\Sigma_{j'=1}^n exp(\eta_o(\mathbf{s}_{i-1},\mathbf{k}_{j'},\mathbf{h}_T))}.
\end{equation}
where $\mathbf{s}_{i-1}$ is the $i-1$-th hidden state in decoder, $\mathbf{h}_T$ is the final hidden state of the input message, and $\eta_o$ is a multilayer perceptron. Compared to the traditional attention in Equation (\ref{eq: attention-weighted sum}) and Equation (\ref{eq: attention-normalize}), topic attention further leverages the final state of the message (i.e., $\mathbf{h}_T$) to weaken the effect of topic words that are irrelevant to the message in generation and highlight the importance of relevant topic words. As a result, the topic vectors $\{\mathbf{o}_i\}_{i=1}^{T'}$ are more correlated to the content of the input message and noise in topic words is controlled in generation. The message attention and the topic attention forms a joint attention mechanism which allows $\mathbf{c}_i$ and $\mathbf{o}_i$ to jointly affect the generation probability.  The advantage of the joint attention is that it makes words in responses not only relevant to the message, but also relevant to the topics of the message.


We define the generation probability $p(y_i)$ as $p(y_i)=p_V(y_i)+p_{K}(y_i)$, where $p_V(y_i)$ and $p_{K}(y_i)$ are defined by 
\begin{equation}
\label{eq: TAJA decoder function}
\begin{aligned}
&p_V(y_i=w)=\left\{
\begin{aligned}
&\frac{1}{Z}e^{\Psi_V(\mathbf{s}_i,\mathbf{y}_{i-1},w)}, &w\in \mathbf{V} \cup \mathbf{K}\\
&0, &w\notin \mathbf{V} \cup \mathbf{K}
\end{aligned}
\right.\\
&p_{K}(y_i=w)=\left\{
\begin{aligned}
&\frac{1}{Z}e^{\Psi_{K}(\mathbf{s}_i,\mathbf{y}_{i-1},\mathbf{c}_i,w)}, &w\in \mathbf{K}\\
&0, &w\notin \mathbf{K}
\end{aligned}
\right.\\
&\mathbf{s}_i=f(y_{i-1},\mathbf{s}_{i-1},\mathbf{c}_i,\mathbf{o}_i).\\
\end{aligned}
\end{equation}
In Equation (\ref{eq: TAJA decoder function}), $\mathbf{V}$ is a response vocabulary, 
and $f$ is a GRU unit. $\Psi_V(\mathbf{s}_i,y_{i-1})$ and $\Psi_{K}(\mathbf{s}_i,y_{i-1},c_i)$ are defined by
\begin{equation}
\label{TA-Seq2Seq decoder function}
\begin{aligned}
&\Psi_V(\mathbf{s}_i,y_{i-1},w)=\sigma(\mathbf{w}^T(\mathbf{W}_V^s\cdot \mathbf{s}_i+\mathbf{W}_V^y\cdot y_{i-1}+\mathbf{b}_V)),\\
&\Psi_{K}(\mathbf{s}_i,y_{i-1},\mathbf{c}_i,w)= \sigma(\mathbf{w}^T(\mathbf{W}_{K}^s\cdot\mathbf{s}_i+\mathbf{W}_{K}^y\cdot y_{i-1}\\
&\indent\indent\indent\indent\indent\indent\indent\indent\indent+\mathbf{W}_{K}^c\cdot \mathbf{c}_{i}+\mathbf{b}_{K})).
\end{aligned}
\end{equation}
where $\sigma(\cdot)$ is tanh, $\mathbf{w}$ is a one-hot indicator vector of word $w$, and $\mathbf{W}_V^s$, $\mathbf{W}_{K}^s$, $\mathbf{W}_V^y$, $\mathbf{W}_{K}^y$, $\mathbf{b}_V$, and 
$\mathbf{b}_{K}$ are parameters.  $Z=\Sigma_{v\in \mathbf{V}}e^{\Psi_V(\mathbf{s}_i,\mathbf{y}_{i-1},v)}+\Sigma_{v'\in \mathbf{K}}e^{\Psi_{K}(\mathbf{s}_i,\mathbf{y}_{i-1},\mathbf{c}_i,v')}$ is a normalizer.

Equation (\ref{eq: TAJA decoder function}) means that the generation probability in TA-Seq2Seq is biased to topic words. For non topic words, the probability (i.e., $p_V(y_i)$) is similar to that in sequence-to-sequence model but with the joint attention mechanism. For topic words, there is an extra probability item $p_{K}(y_i)$ that biases the overall distribution and further increases the possibility of the topic words appearing in responses. The extra probability is determined by the current hidden state of the decoder $\mathbf{s}_i$, the previous word in generation $y_{i-1}$, and the context vector $\mathbf{c}_i$. It means that given the generated parts and the input message, the more relevant a topic word is, the more possible it will appear in the response.  

An extra advantage of TA-Seq2Seq is that it makes better choice on the first word in response generation. The first word matters much because it is the starting point of the language model of the decoder and plays a key role in making the whole response fluent. If the first word is wrongly chosen, then the sentence may never have a chance to go back to a proper response. 
In Seq2Seq with attention, the generation of the first word is totally determined by $\mathbf{c}_0$ which only depends on $\{\mathbf{h}_t\}_{t=1}^T$ since there is no $\mathbf{s}_{i-1}$ when $i=0$. While in TA-Seq2Seq, the first word is generated not only by $\mathbf{c}_0$, but also by $\mathbf{o}_0$ which consists of topic information.  Topic information can help calibrate the selection of the first word to make it more accurate.

We conduct topic learning and response generation in two separate steps rather than let them deeply coupled like VHRED \cite{serban2016hierarchical}. By this means we can leverage extra data from various sources (e.g., web and knowledge base) in response generation. For example, in this work, we estimate topic words from posts in Sina Weibo and provide extra topic information for message-response pairs.

We also encourage the appearance of topic words in responses in a very natural and flexible way by biasing the generation distribution. Through this method, our model allows appearance of multiple topic words rather than merely fixing a single key word in responses like what Mou et al. did in their work \cite{DBLP:journals/corr/MouSYL0J16}. 

\section{Experiments}

We compare TA-Seq2Seq with the-state-of-the-art response generation models by both automatic evaluation and human judgment. 

\subsection{Experiment setup}
We built a data set from Baidu Tieba which is the largest Chinese forum allowing users to post and comment to others' posts. We crawled $20$ million post-comment pairs and used them to simulate message-response pairs in conversation. We removed pairs appearing more than $50$ times to prevent them from dominating learning, and employed Stanford Chinese word segmenter\footnote{\url{http://nlp.stanford.edu/software/segmenter.shtml}} to tokenize the remaining pairs. Pairs with a message or a response having more than $50$ words were also removed.  After these preprocessing, there were $15,209,588$ pairs left. From them, we randomly sampled $5$ million distinct message-response pairs\footnote{Any two pairs are different on messages or responses.} as training data, $10,000$ distinct pairs as validation data, and $1,000$ distinct messages with their responses as test data. Messages in the test pairs were used to generate responses, and responses in the test pairs were treated as ground truth to calculate the perplexity of generation models. There is no overlap among messages in training, validation, and test. We kept $30,000$ most frequent words in messages in the training data to construct a message vocabulary. The message vocabulary covers $98.8\%$ words appearing in messages. Similarly, we constructed a response vocabulary that contains $30,000$ most frequent words in responses in the training data and covers $98.3\%$ words in responses.

We crawled $30$ million posts from Sina Weibo to train a Twitter LDA model. We set the number of topics $T$ as $200$ and the hyperparameters of Twitter LDA as $\alpha=1/T$, $\beta=0.01$, $\gamma=0.01$. For each topic, we selected top $100$ words as topic words.  To filter out universal words, we calculated word frequency using the $30$ million posts, and removed $2000$ words with the highest frequency from the topic words. Words outside the topic words, the message vocabulary, and the response vocabulary were treated as ``UNK''. 
\begin{table}[ht]\small
\label{tab:human}
\centering
 \begin{tabular}{|c|c|c|c|c|}
 \hline
Models & +2 & +1 & 0 & Kappa\\
\hline
S2SA &32.3\%&36.7\%&31.0\%&0.8116\\
\hline
S2SA-MMI&33.1\%&34.8\%&32.1\%&0.7848\\
\hline
S2SA-TopicConcat&35.9\%&29.3\%&34.8\%&0.6633\\
\hline
S2SA-TopicAttention&42.3\%&27.6\%&30.0\%&0.8299\\
\hline
TA-Seq2Seq&44.7\%&24.9\%&30.4\%&0.8417\\
\hline
\end{tabular}
\caption{Human annotation results}
\end{table}
\begin{figure*}[t]
\begin{center}
\includegraphics[width=18cm,height=4.5cm]{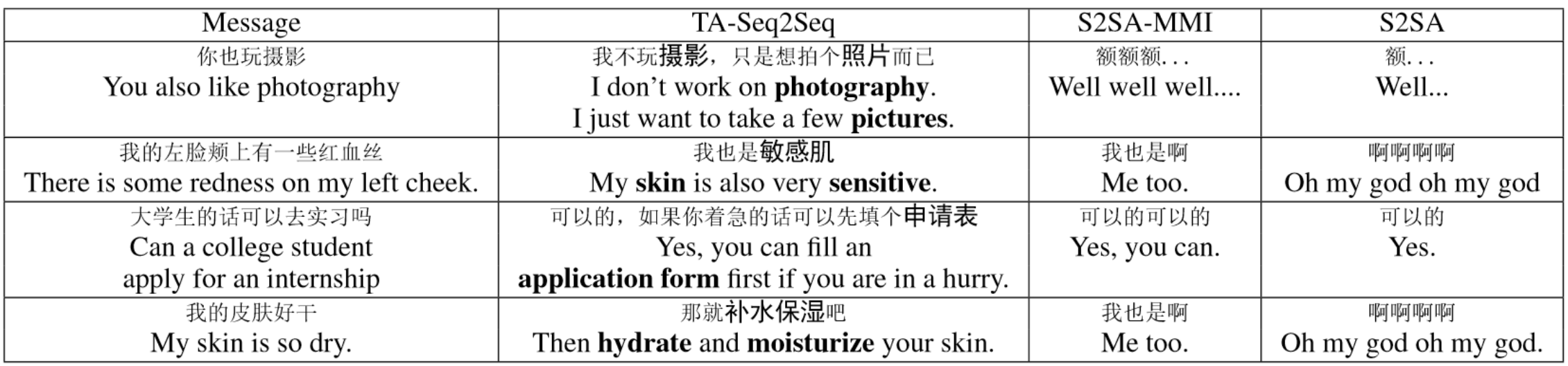}
\end{center}
\caption{Case study}\label{fig:topic_model}
\end{figure*}
\subsection{Evaluation metrics}
How to evaluate a response generation model is still an open problem but not the focus of the paper. Therefore, we followed the existing work and employed the following metrics: 

\textbf{Perplexity}: following \cite{DBLP:journals/corr/VinyalsL15} and \cite{mikolov2010recurrent}, we employed perplexity as an evaluation metric.  Perplexity is defined by Equation (\ref{perplexity}). It measures how well the model predicts a response. A lower perplexity score indicates better generation performance. In this work, perplexity on validation (PPL-D in Table 2) was used to determine when to stop training. If the perplexity stops decreasing and the difference is smaller than $2.0$ five times in validation, we think that the algorithm has reached its convergence and terminate training. We tested the generation ability of different models by perplexity on the test data (PPL-T in Table 2).    
\begin{equation}
\label{perplexity}
PPL=exp\left\{-\frac{1}{N}\Sigma_{i=1}^{N}\log(p(\mathbf{Y_i}))\right\}.
\end{equation}

\textbf{Distinct-1 \& distinct-2}: we counted numbers of distinct unigrams and bigrams in the generated responses. We also followed \cite{li2015diversity} and divided the numbers by total numbers of unigrams and bigrams. We denoted the metrics (both the numbers and the ratios) as distinct-1 and distinct-2 respectively. The two metrics measure how informative and diverse the generated responses are. High numbers and high ratios mean that there is much content in the generated responses, and high numbers further indicate that the generated responses are long.    

\textbf{Human annotation}: in addition to the automatic metrics above, we further recruited human annotators to judge the quality of the generated responses of different models.  Three labelers with rich Tieba experience were invited to do evaluation. Responses generated by different models (the top one response in beam search) were pooled and randomly shuffled for each labeler. Labelers referred to the test messages and judged the quality of the responses according to the following criteria:

\textbf{+2}: The response is not only relevant and natural, but also informative and interesting.

\textbf{+1}: The response can be used as a reply to the message, but it is too universal like ``Yes, I see'' , ``Me too'' and ``I don't know''.

\textbf{0}: The response cannot be used as a reply to the message. It is either semantically irrelevant or disfluent (e.g., with grammatical errors or UNK). 

Agreements among labelers were calculated with Fleiss' kappa \cite{fleiss1973equivalence}. 

Note that we did not choose BLEU \cite{papineni2002bleu} as an evaluation metric, because it has been proven by Liu et al. \cite{liu2016not} that BLEU is not a proper metric for evaluating conversation models as there is weak correlation between BLEU and human judgment.
\begin{table}[ht]\footnotesize
\label{tab:metric}
\centering
 \begin{tabular}{|c|c|c|c|c|}
 \hline
Models &PPL-D&PPL-T&distinct-1& distinct-2\\
\hline
S2SA &147.04&133.11&604/.091&1168/.207 \\
\hline
S2SA-MMI&147.04&133.11&603/.151&1073/.378\\
\hline
S2SA-TopicConcat&150.45&132.12&898/.116&2197/.327\\
\hline
S2SA-TopicAttention&133.81&\textbf{119.55}&894/.106&2057/.277\\
\hline
TA-Seq2Seq&134.63&\textbf{122.82}&1355/.161&2970/.401\\
\hline
\end{tabular}
\caption{Results on automatic metrics}
\end{table}

\subsection{Baselines}
We considered the following baselines.  

\textbf{S2SA}: the standard Seq2Seq model with attention. 

\textbf{S2SA-MMI}: the best performing model in \cite{li2015diversity}. 

\textbf{S2SA-TopicConcat}: to verify the effectiveness of the topic attention of TA-Seq2Seq, we replaced $\mathbf{o}_i$ given by the topic attention in $\mathbf{s}_i$ in Equation (\ref{eq: TAJA decoder function}) by a simple topic vector. The simple topic vector is obtained by concatenating embeddings of topic words and transforming the concatenation to a vector that has the same dimension with the context vector by an MLP. 

\textbf{S2SA-TopicAttention}: to verify the effectiveness of biased generation probability of TA-Seq2Seq, we kept the topic attention but removed the bias probability item which is specially designed for topic words from the generation probability in Equation (\ref{eq: TAJA decoder function}). 

Note that S2SA-TopicConcat and S2SA-TopicAttention are variants of our TA-Seq2Seq. 

In all models, we set the dimensions of the hidden states of the encoder and the decoder as $1000$, and the dimensions of  word embeddings as $620$. All models were initialized with isotropic Gaussian distributions $\mathcal{X} \sim \mathcal{N}(0,0.01)$ and trained with an AdaDelta algorithm \cite{zeiler2012adadelta} on a NVIDIA Tesla K40 GPU. The batch size is $128$. We set the initial learning rate as 1.0 and reduced it by half if the perplexity on validation began to increase. We implemented the models with an open source deep learning tool Blocks\footnote{\url{https://github.com/mila-udem/blocks}}, and shared the code of our model at \url{https://github.com/LynetteXing1991}.

\subsection{Evaluation Results}
Table 1 shows the human annotation results. It is clear that topic aware models (S2SA-TopicConcat, S2SA-TopicAttention and TA-Seq2Seq) generate much more informative and interesting responses (responses labeled as ``+2'') and much less universal responses than the baseline models (S2SA and S2SA-MMI). Among them, TA-Seq2Seq achieves the best performance. Compared with S2SA-MMI, it increases $11.6\%$ ``+2'' responses and reduces $9.9\%$ ``+1'' responses.  S2SA-TopicAttention performs better than S2SA-TopicConcat, meaning that the joint attention mechanism contributes more to response quality than the biased probability in generation. All models
have a proportion of unsuitable responses (labeled as ``0'') around $30\%$ but S2SA-TopicConcat and S2SA-MMI generate more bad responses. This is because without joint attention, noise in topics is brought to generation by the concatenation of topic word embeddings in S2SA-TopicConcat, and in S2SA-MMI, both good responses and bad responses are boosted in re-ranking.  All models have high kappa scores, indicating that labelers reached high agreement regarding to the quality of responses.  We also conducted sign test between TA-Seq2Seq and the baseline models and results show that the improvement from our model is statistically significant ($p$-value $<0.01$).

Table 2 gives the results of automatic metrics. TA-Seq2Seq and S2SA-TopicAttention achieve comparable perplexity on validation data and test data, and both of them are better than the baseline models. We conducted t-test on PPL-T and the results show that the improvement is statistically significant ($p$-value $<0.01$). On distinct-1 and distinct-2, all topic aware models perform better than the baseline models in terms of numbers of distinct n-grams (n=1,2). Among them, TA-Seq2Seq achieves the best performance in terms of both the absolute numbers and the ratios. The results further verified our claim that topic information is helpful on enriching the content of responses. Note that TopicConcat and TopicAttention are worse than S2SA-MMI on ratios of distinct n-grams. This is because responses from S2SA-MMI are generally shorter than those from TopicConcat and TopicAttention. The perplexities of S2SA and S2SA-MMI are the same because S2SA-MMI is an after-processing mechanism on the responses generated by S2SA. Thus we report the perplexity of S2SA to approximately represent the generation ability of S2SA-MMI.


\subsection{Case study}
Figure 3 compares TA-Seq2Seq with S2SA-MMI and S2SA using some examples. Topic words in the responses from TA-Seq2Seq are bolded. From the comparison,  we can see that in TA-Seq2Seq, topic words not only help form the structure of responses, but also act as ``building blocks'' and lead to responses that carry rich information. For example, in case 2, topic information provides prior knowledge to  generation that redness on skin is usually caused by sensitivity of skin and helps form a targeted and informative response. On the other hand, although responses from S2SA-MMI and S2SA also echoed the message, they carry little information and easily lead the conversation to an end.

\section{Related work}
Based on the sequence-to-sequence framework, many generation models have been proposed to improve the quality of generated responses from different perspectives. For example, A. Sordoni et al. \cite{Alessandro2015A} represented the utterances in previous turns as a context vector and incorporated the context vector into response generation. Li et al. \cite{Jiwei2016A} tried to build a personalized conversation engine by adding personal information as extra input. Gu et al. \cite{Jiatao2016Incorporating} introduced copynet to simulate the repeating behavior of human in conversation. Yao et al. \cite{Yao2015Attention} added an extra RNN between the encoder and the decoder of the sequence-to-sequence model with attention to represent intentions. In this work, we consider incorporating topic information into the sequence-to-sequence model. Similar to Li et al. \cite{li2015diversity}, we also try to avoid safe responses in generation. The difference is that we solve the problem by actively bringing content into responses through topics and enriching information carried by the generated responses.  

\section{Conclusion}
We propose a topic aware sequence-to-sequence (TA-Seq2Seq) model to incorporate topic information into response generation. The model leverages the topic information by a joint attention mechanism and a biased generation probability. Empirical study on both automatic evaluation metrics and human annotations shows that the model can generate informative and diverse responses and significantly outperform the-state-of-the-art generation models.

\bibliography{formatting-instructions-latex}
\bibliographystyle{aaai}

\end{document}